\documentclass[runningheads]{llncs}
\usepackage{graphicx}
\usepackage{multirow}
\usepackage{float}
\usepackage{booktabs}
\usepackage{pgfplots}
\pgfplotsset{compat=1.16}
\usepackage{amsmath}
\usepackage{amsfonts} 
\usepackage{dsfont}

\DeclareMathOperator*{\argmax}{arg\,max} 
\newcommand{\ourModel}{CICA} 

\begin{document}

\title{CICA: Content-Injected Contrastive Alignment for Zero-Shot Document Image Classification}
\titlerunning{CICA}
\author{Sankalp Sinha \inst{1,2}\orcidID{0009-0009-6820-3633} \and
Muhammad Saif Ullah Khan \inst{1,2}\orcidID{0000-0002-7375-807X} \and
Talha Uddin Sheikh \inst{1,2}\orcidID{0009-0004-9156-5679} \and
Didier Stricker\inst{1,2} \and
Muhammad Zeshan Afzal \inst{1,2}\orcidID{0000-0002-0536-6867}
}
\authorrunning{Sinha et al.}

\institute{Department of Computer Science, RPTU, 67663 Kaiserslautern, Germany \and
German Research Center for Artificial Intelligence (DFKI), 67663 Kaiserslautern, Germany
\\
\email{\{sankalp.sinha, muhammad\_saif\_ullah.khan, talha\_uddin.sheikh, didier.stricker, muhammad\_zeshan.afzal\}@dfki.de}
}
\maketitle 

\begin{abstract}

Zero-shot learning has been extensively investigated in the broader field of visual recognition, attracting significant interest recently. However, the current work on zero-shot learning in document image classification remains scarce. The existing studies either focus exclusively on zero-shot inference, or their evaluation does not align with the established criteria of zero-shot evaluation in the visual recognition domain. We provide a comprehensive document image classification analysis in Zero-Shot Learning (ZSL) and Generalized Zero-Shot Learning (GZSL) settings to address this gap. Our methodology and evaluation align with the established practices of this domain. Additionally, we propose zero-shot splits for the RVL-CDIP dataset. Furthermore, we introduce \ourModel \footnote{pronounced ``ki·ka''.}, a framework that enhances the zero-shot learning capabilities of CLIP. \ourModel~consists of a novel 'content module' designed to leverage any generic document-related textual information. The discriminative features extracted by this module are aligned with CLIP's text and image features using a novel 'coupled-contrastive' loss. Our module improves CLIP's ZSL top-1 accuracy by 6.7\% and GZSL harmonic mean by 24\% on the RVL-CDIP dataset. Our module is lightweight and adds only 3.3\% more parameters to CLIP. Our work sets the direction for future research in zero-shot document classification.

\end{abstract}

\section{Introduction}
Document image classification is the task of categorizing document images into predefined classes based on their visual and/or textual content~\cite{liu2021document}. This process enhances efficiency by automating the organization of vast quantities of digitized documents~\cite{chen2007survey,liu2021document}. It significantly contributes to downstream processes like information extraction by making it easier to extract valuable insights from documents~\cite{chen2007survey}. Moreover, document image classification is vital for ensuring compliance and effective record-keeping across several industries, including law, healthcare, and finance~\cite{chen2007survey,liu2021document}. In today's digital era, where the volume of digitized documents is rapidly increasing, automated classification is an essential tool to manage this large data effectively~\cite{chen2007survey,liu2021document}.

Traditional approaches have treated document image classification as a supervised learning problem~\cite{afzal2015deepdocclassifier,afzal2017cutting,kang2014convolutional,harley2015evaluation,tensmeyer2017analysis}. In supervised learning, the model learns to map input data to the corresponding labels, focusing on features directly relevant to accomplish this task~\cite {liu2021document}. This requires extensive labeled data for training. Recent advancements in visual recognition tasks have moved towards self-supervised learning~\cite{chen2020simple,grill2020bootstrap} to mitigate this reliance on extensive labeling. Self-supervised pretraining alleviates the burden of feature extraction by utilizing auxiliary pretext tasks, which facilitate the network in comprehending the general structure of images~\cite{gui2023survey}. After this pretraining phase, the model undergoes fine-tuning in a supervised manner with the available labeled dataset. A novel approach in this domain involves integrating these self-supervised pretext tasks to enhance the efficacy of document image classification systems~\cite{cosma2020self}.

Methods utilizing self-supervised and transfer learning significantly reduce dependence on labeled data. Nonetheless, these methods encounter challenges with domain shifts and exhibit limited zero-shot learning capabilities. This limitation is particularly evident when dynamically modifying the number of classes during runtime~\cite{khalifa-etal-2023-contrastive,siddiqui2021analyzing}.

In zero-shot learning, models trained on a set of seen classes are asked to predict classes not in training data. CLIP (Contrastive Language–Image Pretraining)~\cite{radford2021learning} represents a significant leap in zero-shot transfer capabilities in the visual recognition domain. It demonstrated impressive performance on multiple benchmarks by aligning images with textual descriptions. CLIP utilizes a large image-text pair dataset, enabling the model to extensively learn various visual concepts and their associations with natural language~\cite{radford2021learning}.

The field of zero-shot learning for document images is still emerging. However, the zero-shot proficiency of CLIP has sparked some notable research, such as the study by Khalifa et al.~\cite{khalifa-etal-2023-contrastive}, demonstrating the enhancement of zero-shot classification through contrastive training in unstructured documents. Additionally, the analysis by Siddiqui et al.~\cite{siddiqui2021analyzing} offers a critical examination of CLIP's zero-shot transfer capabilities using the RVL-CDIP dataset~\cite{harley2015icdar}.

The current work on zero-shot learning in document images remains scarce~\cite{siddiqui2021analyzing,khalifa-etal-2023-contrastive}, with existing studies lacking a unified evaluation framework. This absence of a standardized evaluation criterion is particularly notable, considering the well-established evaluation practices in the broader field of visual recognition, as meticulously outlined by Xian et al.~\cite{xian2018zero}. Xian et al.~\cite{xian2018zero} propose a unified evaluation framework that emphasizes average per-class top-1 accuracy as a critical metric to ensure fairness across datasets, particularly those with imbalanced class distributions, in the zero-shot learning (ZSL) scenario. Additionally, it introduces the harmonic mean of seen and unseen accuracies as a balanced evaluation metric for generalized zero-shot learning (GZSL) scenarios. To the best of our knowledge, zero-shot learning has not been comprehensively implemented within the document image domain. This area continues to be under-researched and represents a significant gap in current literature.

{\textbf{In this paper}, we address this research gap of zero-shot learning in document image classification. To achieve this, we propose a novel 'content module' specifically designed to process general information from documents. In our case, this information consists of text extracted from document images using OCR. Moreover, we present a unique 'coupled contrastive' loss mechanism. This mechanism is designed to align the features from our 'content module' with both the text and image features of the CLIP model, adhering to CLIP's contrastive alignment principles. These innovations together form a cohesive, end-to-end trainable framework called \ourModel~that significantly improves the zero-shot learning capabilities of CLIP. This framework is meticulously evaluated under the standardized criterion outlined in~\cite{xian2018zero}. Our code is publicly available\footnote{\url{https://github.com/mindgarage/cica}}. The primary contributions of our research are outlined as follows:

\begin{itemize}
\renewcommand\labelitemi{$\bullet$} 
    \item This work represents, to the best of our knowledge, the first instance of implementing Zero-Shot Learning (ZSL) and Generalized Zero-Shot Learning (GZSL) within the domain of document image classification.
    \item We introduce a novel 'content module' that processes generic information from documents. In our context, this information is the OCR-extracted text from document images.
    \item We introduce a novel 'coupled contrastive loss' that aligns the features of our 'content module' with the text and image features of CLIP.
    \item We propose novel ZSL and GZSL splits for the RVL-CDIP dataset, grounded on CLIP’s performance, and conduct a quantitative analysis of CLIP’s zero-shot transfer potential.
    \item We conduct comprehensive experiments and ablation studies within the ZSL and GZSL settings to rigorously evaluate our \ourModel~framework. 
\end{itemize}

\section{Related Work}
\subsection{Document Image Classification}

Supervised deep learning approaches have significantly outperformed rule-based methods and hand-crafted features in document image classification \cite{kang2014convolutional,afzal2015deepdocclassifier,harley2015evaluation,tensmeyer2017analysis,afzal2017cutting,kolsch2017real,kanchi2022emmdocclassifier}, primarily due to their ability to learn complex patterns and representations directly from data. Unlike rule-based systems \cite{lowe2004distinctive,ke2004pca,bay2006surf,mikolajczyk2005performance,kumar2012learning} that rely on predefined rules and manual feature engineering, deep learning models automatically extract and learn the most relevant features for a given task. However, the effectiveness of deep learning models heavily depends on the availability of large training data. This requirement can be challenging, as obtaining and labeling a vast and varied dataset of document images is time-consuming and resource-intensive \cite{liu2021document}.

In the field of document image classification, another prevalent approach involves leveraging transfer learning~\cite{das2018document,afzal2015deepdocclassifier} specifically fine-tuning to attain competitive results. This method involves pre-training a model on a substantial image-classification dataset, such as ImageNet~\cite{krizhevsky2012imagenet}, thereby transforming the initial layers of the convolutional network, or potentially even transformer-based models like Vision Transformers (ViT)~\cite{dosovitskiy2020image} into universal feature extractors. Subsequently, the model undergoes fine-tuning, particularly after reinitializing the output layer to suit the specific dataset.

Recent trends in visual recognition have shifted towards semi-supervised and self-supervised learning to reduce reliance on extensive labeled datasets. In SimCLR~\cite{chen2020simple}, Chen et al. introduce a contrastive learning-based self-supervised framework to distinguish similar and dissimilar image pairs, enhancing learning without labeled data. Grill et al. present BYOL~\cite{grill2020bootstrap}, which avoids negative pairs in contrastive learning and uses a dual neural network for learning from image augmentations. Following these advancements, Cosma et al.~\cite{cosma2020self} introduce multimodal self-supervised learning for document images, using techniques such as 'Jigsaw Whole' for effective context-sensitive pretraining and text topic prediction from images in data-scarce scenarios. Dauphinee et al.~\cite{dauphinee2019modular} further the application of multimodal models by employing separate classifiers for visual and textual data, achieving notable accuracy on the RVL-CDIP dataset. These sequential studies highlight the evolving sophistication in multimodal integration for document image classification.

\subsection{Zero-Shot Image Classification}

Zero-Shot Learning (ZSL) is a machine learning paradigm designed to classify objects or entities not present in the training data. In ZSL, a model trained on a set of (seen) classes is evaluated on a different set of previously unseen classes~\cite{xian2018zero}. Generalized zero-shot learning extends ZSL by including both seen and unseen classes in the test phase, making it a more realistic and challenging scenario. GZSL aims to balance the classification performance across both seen and unseen classes, addressing the inherent bias towards seen classes due to the training process~\cite{xian2018zero}.

In the literature, this is accomplished by utilizing supplementary information associated with a unique group of previously unknown classes. Various approaches seek to create a link between image attributes and a class representation that incorporates supplementary information~\cite{changpinyo2016synthesized,mancini2021open,romera2015embarrassingly,xian2016latent}. However, these methods frequently face challenges with biases towards classes that have not been previously observed. To address this issue, a different set of techniques aims to understand the distribution of image features through generative modeling~\cite{brendel2019approximating,verma2018generalized,schonfeld2019generalized}. In this scenario, some methods directly utilize semantic information to produce features for new and previously unknown classes~\cite{schonfeld2019generalized,zhu2018generative}. Certain strategies concentrate on developing a class-conditional generator capable of generating features for classes that have not been seen before~\cite{verma2018generalized,brendel2019approximating}. After being trained with data from known classes, these models employ supplementary information to create features for classes they have not encountered before, thereby mitigating issues of bias \cite{naeem2023i2mvformer}. Additionally, other studies aim to improve the integration of visual and semantic data~\cite{cacheux2019modeling,jiang2019transferable} and to enhance the performance of feature extraction from images through advanced training methods~\cite{xu2020attribute,ji2018stacked}.

CLIP~\cite{radford2021learning}, noteworthy for its zero-shot transfer capabilities, revolutionized computer vision and NLP integration by learning from natural language without relying on labeled data. It used a vast collection of 400 million internet-sourced image-text pairs for training, focusing on matching images to text descriptions. Integrating additional modalities into CLIP to enhance its native performance in zero-shot settings is an emerging research field. Our work draws inspiration from Khan et al.~\cite{khan2024focusclip}, who demonstrate that heatmaps can be used as an additional modality within the CLIP framework to enhance the performance of various human-centric classification tasks.

\subsection{Zero-Shot Document Image Classification}
Zero-shot learning, within the context of document image classification, is an emerging subfield within zero-shot learning that has not been extensively studied. Khalifa et al.~\cite{khalifa-etal-2023-contrastive} demonstrate how contrastive training methods can enhance zero-shot classification performance for unstructured documents. Similarly, the research by Siddiqui et al.~\cite{siddiqui2021analyzing} focuses on investigating the potential of zero-shot learning for document image classification. Their study provides a critical analysis of the zero-shot transfer abilities of the CLIP model. It leverages CLIP's pre-trained weights to evaluate the average top-1 accuracy across the RVL-CDIP dataset, highlighting CLIP's zero-shot potential.

\section{Methodology}\label{Methodology}
Section~\ref{Methodology} offers a comprehensive overview of our proposed~\ourModel~framework. In Subsection~\ref{Problem-Formulation}, we explore the problem formulation for zero-shot learning. Subsection~\ref{Model-Architecture} introduces our innovative 'content module' and its integration within the CLIP model. Subsection~\ref{Loss-Function} details the formulation of our proposed novel 'coupled contrastive' loss. Lastly, Subsection~\ref{Inference} outlines the procedures for model inference. Together, these subsections systematically develop the narrative of the end-to-end trainable \ourModel~framework, covering everything from its theoretical basis to its training and inference mechanisms.
\subsection{Problem Formulation}\label{Problem-Formulation}


Zero-shot learning aims to enhance a model's ability to generalize to classes not present in its initial training. This is achieved by leveraging auxiliary information related to a distinct set of unknown classes. In this context, we define the set of classes in the training dataset as \( Y_s \) and the set of classes only encountered during testing as \( Y_u \). Hence, \( Y_s \) is the set of seen classes, and \( Y_u \) is the set of unseen classes. 
Consider our training dataset \(T = \{(x, y, c) | x \in X_s, y \in Y_s, C \in C_s\}\). In this dataset, \(x\) is an RGB image from the training set \(X_s\). The term \(y\) refers to the corresponding label from the seen classes \(Y_s\). Furthermore, \(c\) represents the associated information (content) from the document, where \(C_s\) is the content set for the seen classes. During testing, an additional content set \( C_u \) is also made available, corresponding to the novel, unseen classes. This is defined such that \( C = C_s \cup C_u \) and \( C_s \cap C_u = \emptyset \). In the ZSL scenario, the model has to classify an instance into one of the unseen classes \( Y_u \). In the GZSL scenario, the model's objective is extended to accurately classify instances into both the seen and unseen classes, combined as \( Y = Y_s \cup Y_u \).

\subsection{Model Architecture}\label{Model-Architecture}
The proposed model extends the CLIP architecture by integrating a plug-and-play 'content module' as seen in \textit{Fig. \ref{fig1}}. This module is trained using our proposed 'coupled contrastive' loss, which harmonizes the features of the content encoder with those of the image and text encoders.

\begin{figure}[H]
\includegraphics[angle=-90, width=\textwidth]{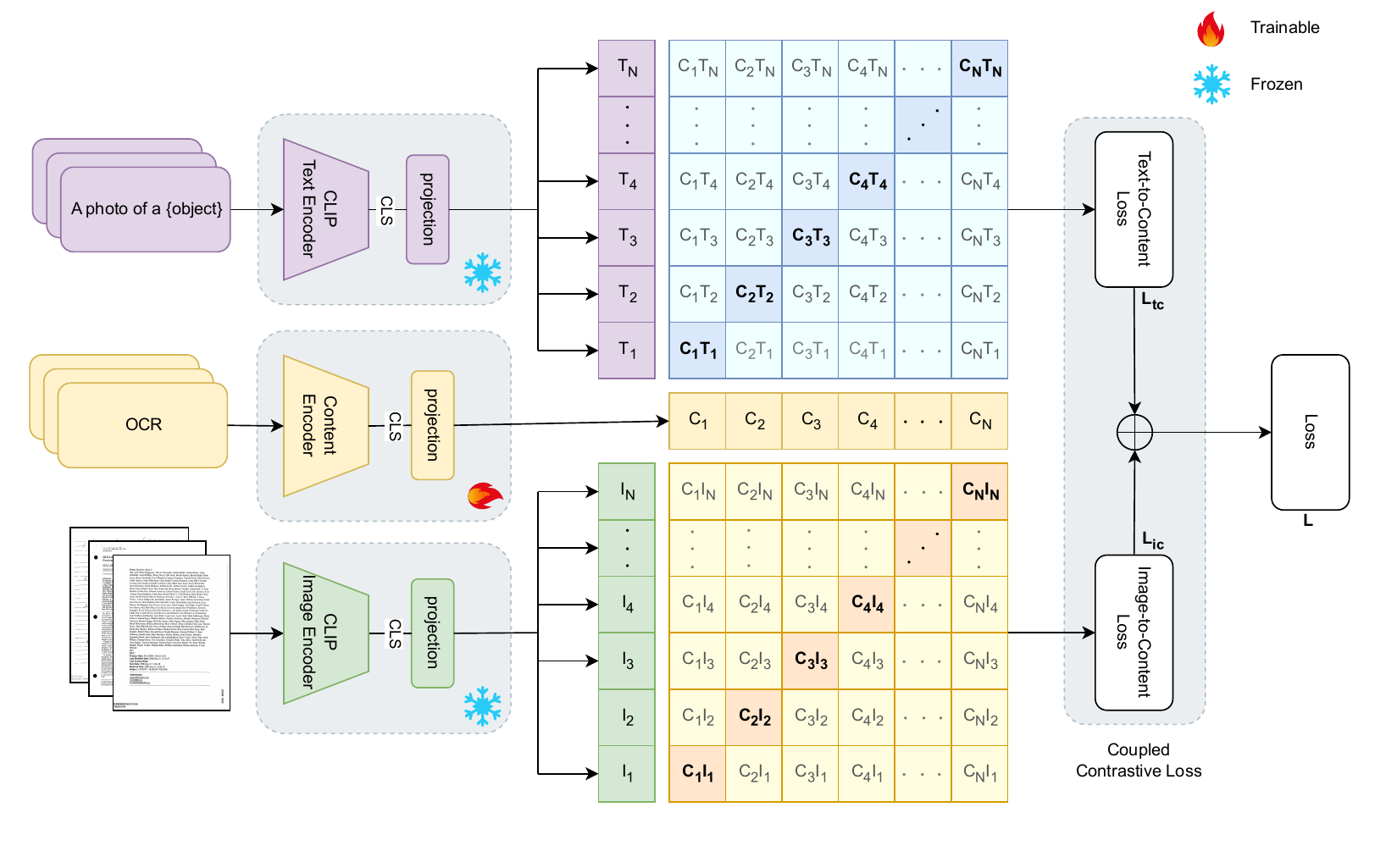}
\caption{Illustration of \ourModel's architecture and training paradigm, highlighting how image and text features from CLIP are integrated with the content module's features through a coupled contrastive loss. This aligns positive image-content and text-content pairs (highlighted in the matrix) and distances negative pairs (lightly shaded in the matrix). $N$ denotes the batch size.}
\label{fig1}
\end{figure}

Our model leverages CLIP's image and text encoders to process both visual and textual data. Our proposed content module includes a transformer encoder block, complete with a dedicated token and position embedding layer.
The 'coupled contrastive' loss introduces an element of self-supervised learning. It promotes a unified and coherent feature space among all encoders in the model, empowering the content encoder to explore and delineate inherent relationships within the data autonomously. This strategy aligns seamlessly with the contrastive pre-training framework employed by CLIP.

\subsection{Loss Function}\label{Loss-Function}
Given a batch of $N$ images $\mathds{I} = \{x_i\}$, texts $\mathds{T} = \{t_i\}$, and contents $\mathds{C} = \{c_i\}$, let $f_i$, $f_t$, and $f_c$ represent the image, text, and content encoders, respectively. Each image \( x_i \in \mathbb{R}^{H \times W \times Ch} \), where \( H \) and \( W \) denote the image's height and width, and \( Ch \) represents the RGB color channels. The images are converted into 2D flattened patches \( p_i \in \mathbb{R}^{N \times (P^2 \cdot Ch)} \), where \( P \) is the patch size. This process results in \( M = \frac{HW}{P^2} \) patches per image. Each text \( t_i \) is tokenized using CLIP's tokenizer, with a context length of 77. This tokenization yields a sequence of tokens denoted by \( t'_i \). Similarly, each content \( c_i \) is tokenized with a context length of 512, producing a sequence denoted by \( c'_i \). For each sequence namely: tokenized text \( t'_i \), tokenized content \( c'_i \), and flattened image patches \( p_i \), a CLS token is prepended. The resulting sequences are denoted as \([CLS]_i^I\), \([CLS]_i^T\), and \([CLS]_i^{C}\) respectively. These sequences are then passed through their respective encoders: $f_i$, $f_t$, and $f_c$. After processing, the CLS tokens are extracted and represented as \(CLS_i^I\), \(CLS_i^T\), and \(CLS_i^{C}\). Finally, these CLS tokens are projected into a joint embedding space of 512 dimensions using projection layers. These layers are denoted as \(\text{Proj}^I\), \(\text{Proj}^T\), and \(\text{Proj}^C\) as seen in Equation \ref{equation_1}. The resulting features are normalized, as detailed in Equations \ref{equation_2}.

\begin{equation}
\hat{I_{i}} = \text{Proj}^I(CLS_i^{I}) \qquad \hat{T_{i}} = \text{Proj}^T(CLS_i^{T}) \qquad \hat{C_{i}} = \text{Proj}^C(CLS_i^{C})
\label{equation_1}
\end{equation}

\begin{equation}
I_{i} = \frac{\hat{I_i}}{\|\hat{I_i\|}} \qquad T_{i} = \frac{\hat{T_i}}{\|\hat{T_i\|}} \qquad C_{i} = \frac{\hat{C_i}}{\|\hat{C_i\|}}
\label{equation_2}
\end{equation}

\begin{equation}
L_{i, ic} = -\log \frac{\exp(\phi(I_{i}, C_{i}) / \tau_{IC})}{\sum_{j=1}^{N} \mathds{1}_{[j \neq i]}  \exp(\phi(I_{i}, C_{j}) / \tau_{IC})}
\label{equation_3}
\end{equation}

\begin{equation}
L_{i, tc} = -\log \frac{\exp(\phi(T_{i}, C_{i}) / \tau_{TC})}{\sum_{j=1}^{N} \mathds{1}_{[j \neq i]}  \exp(\phi(T_{i}, C_{j}) / \tau_{TC})}
\label{equation_4}
\end{equation}
Eq.~\ref{equation_3} and Eq.~\ref{equation_4} denote the image-to-content and text-to-content loss components, where $\tau_{IC}$ and $\tau_{TC}$ are the learned temperature parameters for each component. Furthermore, $\phi$ represents cosine similarity, and $\mathds{1}$ is an indicator function.
Eq.~\ref{equation_5} represents the total loss as the average contrastive loss for each image-text-content triplet in the batch, combining the losses from both image-to-content and text-to-content pairs.
\begin{equation}
L = \sum_{i=1}^{N} \frac{1}{2} (L_{i, ic} + L_{i, tc})
\label{equation_5}
\end{equation}


\subsection{Inference}\label{Inference}
\begin{figure}[H]
\includegraphics[angle=-90, width=\textwidth]{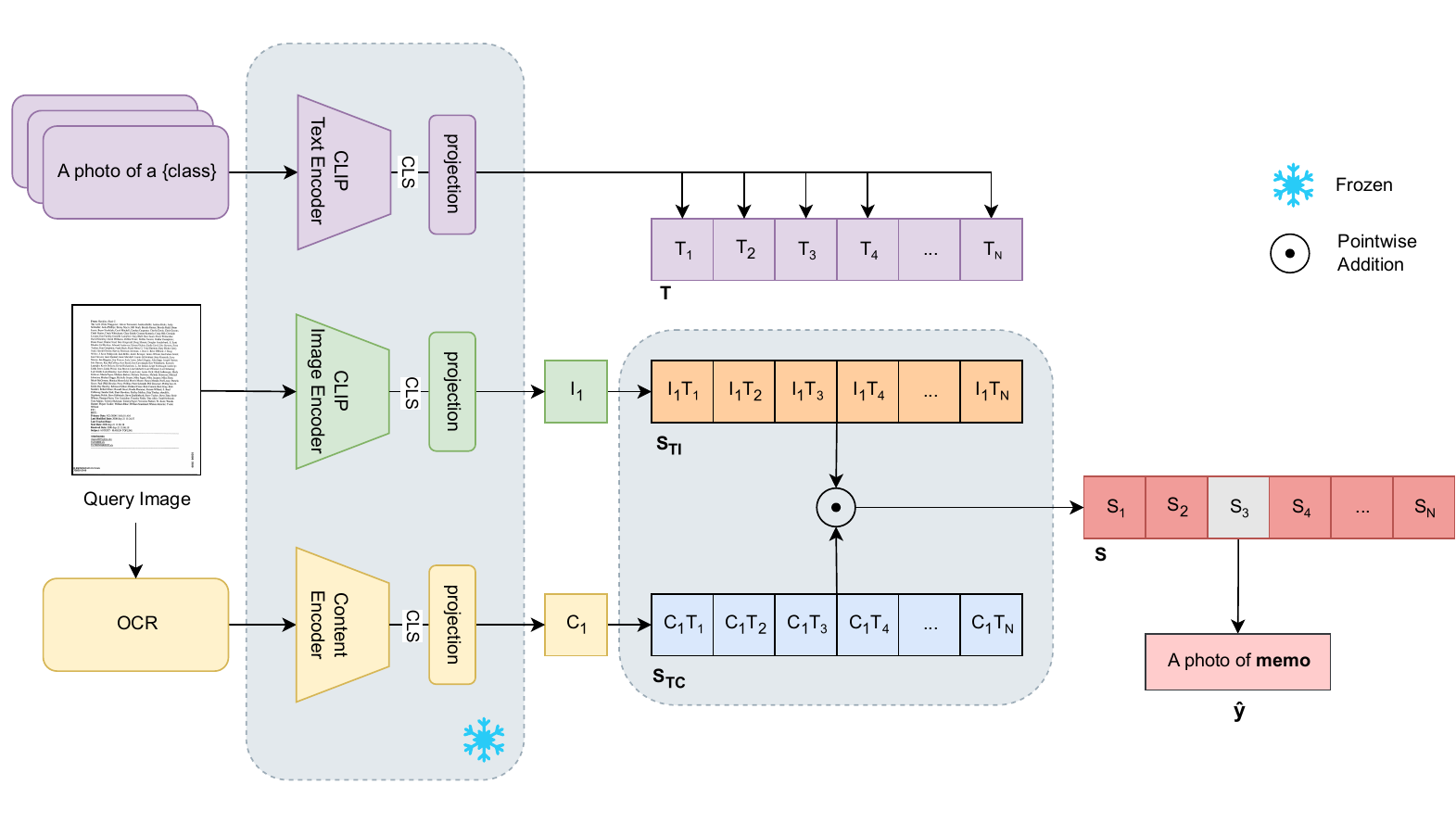}
\caption{Illustration showing the inference logic for~\ourModel. At test time, the learned content encoder, along with the CLIP text encoder, synthesizes a zero-shot linear classifier by embedding a prompt with the names of the test set's classes along with the content for the test sample. Here, $N$ is the number of classes.}
\label{Content-CLIP-Inference}
\end{figure}

Given a test image \( x_i \in \mathds{I} \) and the content for the test image \( c_i \in \mathds{C} \). Let \( \{t_i\} \) be the set of texts for each class.
Using Eq.~\ref{equation_2}, we get the normalized feature \( I_1 \) and \( C_1 \)  for the image and the content, along with the set of normalized features for the texts denoted as \( T = \{T_{i}\} \) where \( i \in [1, N] \) and $N$ is the number of classes.
Now, Eq.~\ref{inference_equation_1} and Eq.~\ref{inference_equation_2} represent the text-to-image and text-to-content logits. The final logits and prediction are represented in Eq.~\ref{inference_equation_3} and Eq.~\ref{inference_equation_4}, respectively.

\begin{equation}
S_{TI} = \tau_{IC} \times I_{1}^T \cdot T
\label{inference_equation_1}
\end{equation}
\begin{equation}
S_{TC} = \tau_{TC}  \times C_{1}^T \cdot T
\label{inference_equation_2}
\end{equation}
\begin{equation}
S = \frac{S_{TI} + S_{TC}}{2}
\label{inference_equation_3}
\end{equation}
\begin{equation}
\hat{y} = \argmax{S}
\label{inference_equation_4}
\end{equation}

\section{Experimental Setup}
This section details the experimental setup for evaluating the proposed~\ourModel~framework. In Subsection~\ref{Datasets}, we describe the dataset utilized. In Subsection~\ref{Evaluation_Criteria}, we outline the evaluation criterion. In Subsection~\ref{Proposed_Splits}, we present the proposed zero-shot data splits. Finally, in Subsection~\ref{Implementation_Details}, we provide implementation details of the model.

\subsection{Datasets}\label{Datasets}

\subsubsection{RVL-CDIP}
The RVL-CDIP dataset~\cite{harley2015icdar} comprises 400,000 grayscale images from a broader collection of 1 million images. It is organized into 16 classes: letters, forms, emails, handwritten notes, advertisements, scientific reports, publications, specifications, file folders, news articles, budgets, invoices, presentations, questionnaires, resumes, and memos. Its wide variety and substantial volume make it a valuable resource for training and evaluating machine learning models. It is especially instrumental for document classification and OCR tasks.


\subsection{Evaluation Criteria}\label{Evaluation_Criteria}
We follow the ZSL and GZSL protocols proposed by Xian et al.~\cite{xian2018zero}. Single-label image classification accuracy is typically assessed using Top-1 accuracy. A prediction is considered accurate if the top predicted class matches the correct one. This accuracy, when averaged across all images, tends to favor densely populated classes. To achieve high performance across both densely and sparsely populated classes, we adjust our method. We start by independently calculating the accuracy of each class. Then, we compute their cumulative sum. This is divided by the total number of classes, effectively measuring the average per-class Top-1 accuracy. This method offers a more balanced evaluation across classes of varying sizes and is used in our ZSL setting~\cite{xian2018zero}.

\begin{equation}
    \text{acc}_Y = \frac{1}{||Y||} \sum_{n=1}^{||Y||} \frac{\text{\# correct predictions in } n}{\text{\# samples in } n}
\end{equation}

The GZSL setting evaluates both test classes \((Y_{u})\) and training classes \((Y_{s})\). This makes the approach more relevant to real-world scenarios. In our testing procedure, we utilize a subset of images from the training classes to calculate the average per-class top-1 accuracy for both the training and test classes. We then compute the harmonic mean of these accuracies, offering a comprehensive measure of the model's performance across seen and unseen classes \cite{xian2018zero}. Additionally, we report the individual accuracies for both seen and unseen classes.

\begin{equation}
    H = 2 \times \frac{\text{acc}_{Y_s} \times \text{acc}_{Y_u}}{\text{acc}_{Y_s} + \text{acc}_{Y_u}}
\end{equation}

Calibrated stacking \cite{chao2016empirical} is widely used in the GZSL setting. We also employ it for our evaluations. Our evaluation methodology involves utilizing a pre-trained and unaltered CLIP model, as described in Radford et al.~\cite{radford2021learning}. We conduct comparative assessments of both CLIP and our proposed \ourModel~model. These assessments occur in ZSL and GZSL settings. We adhere to the evaluation criteria outlined in section \ref{Evaluation_Criteria}. We also utilize the data splits outlined in section \ref{Proposed_Splits}.

\subsection{Splits for Zero-Shot Learning}\label{Proposed_Splits}

We propose two distinct dataset splits for the RVL-CDIP dataset for both the ZSL and GZSL settings. The first is the `sequential splits,' which designate four classes as unseen while keeping the remaining twelve as seen. We create four such distinct splits, covering all 16 dataset classes. This approach guarantees that each class is assigned to the unseen set exactly once. \textit{Tab.~\ref{tab:proposed_splits_sequential}} illustrates these splits.

The second is the `incremental splits,' produced using rank ordering. Given a rank ordering \( R \) of the top-1 class-wise accuracies from the RVL-CDIP dataset in ascending order, obtained using the pre-trained CLIP model. We define the split \( S_i \) as follows:

\[ S^{I}_i = (u,s) \]

where \( s \) represents the set of seen classes and \( u \) represents the set of unseen classes. In this context, the index \( i \) ranges from 2 to 8, indicating the number of classes in the unseen set \( u \). The remaining classes, totaling \( 16 - i \), are included in the seen set \( s \). In accordance with our evaluation criterion in Section \ref{Evaluation_Criteria}, we ensure that $Y_{u} \cap Y_{s} = \emptyset$ and $Y = Y_{u} \cup Y_{s}$. The rank-ordering $R$ for the RVL-CDIP dataset is shown in \textit{Fig. \ref{fig:rank-ordering}}. Essentially, $S^{I}_2$ keeps the top two worst classes in the unseen set and the rest in the seen set. Similarly, $S^{I}_3$ keeps the top three worst performing classes in the unseen set and so on for $S^{I}_4$ to $S^{I}_8$.

\begin{table}[ht]
    \caption{Sequential splits of the RVL-CDIP dataset where four classes are kept in the unseen set, with the rest in the seen set. Table shows the unseen classes for each split.}
    \centering
    \begin{tabular}{cl}
    \toprule
    \textbf{Split} & \textbf{Unseen Classes} \\
    \cmidrule(l){1-1} \cmidrule(l){2-2}
    A & email, form, handwritten, letter \\
    B & advertisement, scientific publication, scientific report, specification \\
    C & budget, file folder, invoice, news article \\
    D & memo, presentation, questionnaire, resume \\
    \bottomrule
    \end{tabular}
    \label{tab:proposed_splits_sequential}
\end{table}

\begin{figure}
    \centering
    \includegraphics[width=\textwidth]{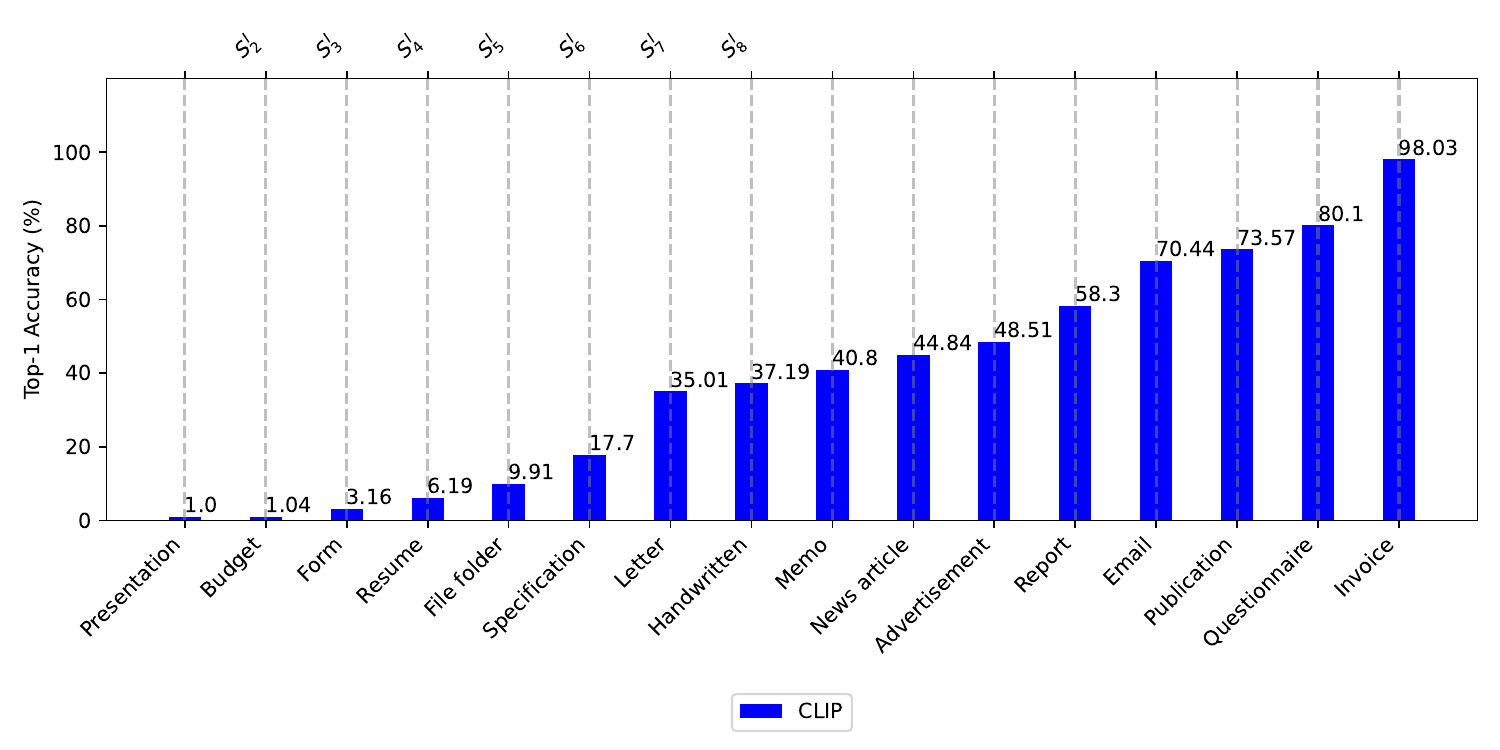}
    \caption{Rank ordering of class-wise top-1 accuracies obtained from the frozen CLIP model, for the RVL-CDIP dataset. The top axis also shows the zone of the varying number of unseen classes in the incremental splits ($S_i^I$).}
    \label{fig:rank-ordering}
\end{figure}

\subsection{Document Information or Content}\label{Metadata}
Our content module is designed to be adaptable, allowing it to process a wide range of additional information about the target classes. However, in our studies, we focus on using text extracted through Optical Character Recognition (OCR) as the content. We employ two OCR engines, Tesseract \cite{TesseractOCR} and docTR \cite{doctr2021}, for this purpose, which are crucial in creating the content set C. The impact of these two OCR engines on our model's performance is analyzed in \textit{Tab.~\ref{OCR-ablation-studies-RVL-CDIP}}.


\subsection{Implementation Details}\label{Implementation_Details}
Our \ourModel~model was developed with PyTorch, mirroring CLIP's data augmentation techniques. We set the learning rate at 1e-3, reducing it to 0.1 after five epochs. The model was trained with a batch size of 32 over 10 epochs. For optimization, we used the Adam optimizer. Unless otherwise specified, all models utilize the OCR as content from the docTR OCR engine. Our model uses the ViT-L/14@336px variant of CLIP as its backbone architecture. All comparisons are against this specific CLIP version. For the finetuned version of CLIP, we use a learning rate at 1e-4, reducing it to 0.1 after five epochs and training for 10 epochs in total. Content for the 400,000 grayscale images in the RVL-CDIP dataset was generated using two OCR engines in $\approx$ 24 hours with 64 CPUs.

\section{Results and Discussion}

\subsection{Results}
\textit{Tab.~\ref{RVL-CDIP-Results-S1-S4}} provides a detailed comparison of the performance between the different models. The table focuses explicitly on the sequential splits of the RVL-CDIP dataset. The results show that our \ourModel~model consistently surpasses CLIP in performance across the splits. While a substantial increase in accuracies for seen classes is observed, it can be noted that simple finetuning of CLIP does not result in performance gains in ZSL and GZSL settings. On the other hand, we observe an average increase of 6.7\% in the top-1 accuracy across the splits for the ZSL setting. Further, in the GZSL setting, we see an average improvement of 24\% in the harmonic mean (\textbf{H}).

\begin{table}[H]
    \centering
    \caption{Comparative performance of CLIP, finetuned CLIP, and \ourModel~across sequential splits (A - D) of the RVL-CDIP dataset, highlighting top-1 accuracy and harmonic mean in \textbf{bold}. \ourModel~consistently outperforms CLIP in both ZSL and GZSL settings on all splits.}
    \begin{tabular}{clcccc}
    \toprule
    & & \multicolumn{4}{c}{\textbf{RVL-CDIP}} \\
    \cmidrule(l){3-6}
    & & \textbf{ZSL} & \multicolumn{3}{c}{\textbf{GZSL}} \\
    \cmidrule(l){3-3} \cmidrule(l){4-6}
    \textbf{Split} & \textbf{Method} & \textbf{T1} & \textbf{u} & \textbf{s} & \textbf{H} \\
    \cmidrule(l){1-1} \cmidrule(l){2-2} \cmidrule(l){3-3} \cmidrule(l){4-4} \cmidrule(l){5-5} \cmidrule(l){6-6}
    \multirow{2}{*}{A}
    & \texttt{CLIP (finetuned)} & 42.62 & 16.50 & 61.60 & 26.02 \\
    & \texttt{CLIP} & 70.55 & 49.74 & 35.61 & 41.50 \\
    & \texttt{\ourModel} & \textbf{70.64} & 61.84 & 69.36 & \textbf{65.38} \\
    \midrule
    \multirow{2}{*}{B}
    & \texttt{CLIP (finetuned)} & 45.73 & 23.73 & 70.92 & 35.56 \\
    & \texttt{CLIP} & 67.02 & 53.83 & 33.88 & 41.58 \\
    & \texttt{\ourModel} & \textbf{77.87} & 69.84 & 75.40 & \textbf{72.52} \\
    \midrule
    \multirow{2}{*}{C}
    & \texttt{CLIP (finetuned)} & 48.06 & 29.38 & 67.98 & 41.03 \\
    & \texttt{CLIP} & 50.35 & 41.61 & 37.31 & 39.34 \\
    & \texttt{\ourModel} & \textbf{59.52} & 53.94 & 72.17 & \textbf{61.74} \\
    \midrule
    \multirow{2}{*}{D}
    & \texttt{CLIP (finetuned)} & 41.42 & 13.25 & 78.99 & 22.69 \\
    & \texttt{CLIP} & 51.67 & 39.37 & 36.29 & 37.77 \\
    & \texttt{\ourModel} & \textbf{61.13} & 51.13 & 71.17 & \textbf{59.51} \\
    \bottomrule
    \end{tabular}
    \label{RVL-CDIP-Results-S1-S4}
\end{table}

These improvements highlight the efficacy of the \ourModel~model. It capitalizes on the inherent capabilities of CLIP's architecture. Additionally, it successfully integrates features tailored to the ZSL and GZSL requirements.

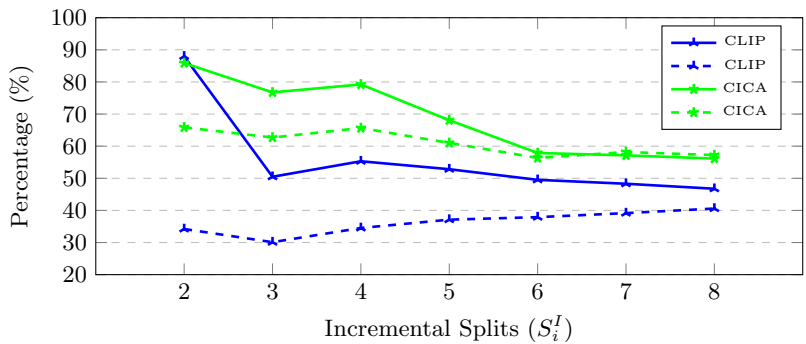
\begin{figure}[H]
    \centering
    \begin{tikzpicture}
    \begin{axis}[
        xlabel={Incremental Splits ($S^{I}_i$)},
        ylabel={Percentage (\%)},
        xmin=1, xmax=9,
        ymin=20, ymax=100,
        xtick={2,3,4,5,6,7,8},
        ytick={20,30,40,50,60,70,80,90,100},
        legend pos=north east,
        legend style={font=\tiny},
        ymajorgrids=true,
        width=0.9\linewidth,
        height=5cm,
        grid style=dashed
    ]
    \addplot[
        color=blue,
        mark=Mercedes star,
        line width=1pt,
        mark options={fill=blue},
        ]
        coordinates {
        (2, 87.96)(3, 50.48)(4, 55.30)(5, 52.83)(6, 49.52)(7, 48.30)(8, 46.75)
        };
        \addlegendentry{CLIP}

    \addplot[
        color=blue,
        mark=Mercedes star,
        line width=1pt,
        style=dashed,
        mark options={fill=blue},
        ]
        coordinates {
        (2, 34.21)(3, 30.11)(4, 34.53)(5, 37.12)(6, 37.90)(7, 39.16)(8, 40.58)
        };
        \addlegendentry{CLIP}

    \addplot[
        color=green,
        mark=star,
        line width=1pt,
        mark options={fill=green},
        ]
        coordinates {
        (2, 85.88)(3, 76.75)(4, 79.23)(5, 68.09)(6, 57.89)(7, 57.14)(8, 56.14)
        };
        \addlegendentry{\ourModel}

    \addplot[
        color=green,
        mark=star,
        line width=1pt,
        mark options={fill=green},
        style=dashed,
        ]
        coordinates {
        (2, 65.87)(3, 62.70)(4, 65.66)(5, 61.01)(6, 56.33)(7, 58.18)(8, 57.24)
        };
        \addlegendentry{\ourModel}
    \end{axis}
    \end{tikzpicture}
    \caption{Comparative analysis of CLIP and \ourModel~on RVL-CDIP incremental splits, showing CICA outperforming CLIP. Dashed lines indicate GZSL outcomes as harmonic mean, while solid lines depict ZSL performance using Top-1 accuracy. We observe that \ourModel~consistently outperforms CLIP.}
    \label{RVL-CDIP-Incremental-Splits}
\end{figure}

Furthermore, \textit{Fig. \ref{RVL-CDIP-Incremental-Splits}} illustrates the performance comparison between CLIP and our proposed \ourModel. The figure specifically focuses on the incremental splits of the RVL-CDIP for both the ZSL and GZSL settings. It is evident that for ZSL scenarios (solid lines), \ourModel~outperforms CLIP in most splits except when $s=2$, where its performance is slightly lower. This can be attributed to the 'presentation' and 'budget' classes being designated as unseen during this split. These classes are significantly different from the rest of the dataset, making it unlikely for our model to infer knowledge about them from the seen classes. In GZSL scenarios (dotted lines), both models demonstrate a decline in performance as the number of unseen classes increases, yet \ourModel~maintains a consistent lead over CLIP. This indicates that \ourModel~has a better capability to generalize to unseen classes. Despite the unique challenge posed by the distinct 'presentation' and 'budget' categories, we see an overall improvement in both the ZSL and GZSL results.

\subsection{Ablation Study}\label{Ablation-Studies}
In our ablation studies, we focus on three crucial aspects that determine our model's effectiveness. Firstly, we analyze the impact of aligning features from our content module specifically with either the image or text features of CLIP. Secondly, we evaluate how the quality of text extracted by different OCR engines influences \ourModel's~accuracy. Thirdly, we investigate the effects of feature fusion techniques on the \ourModel's~ability to predict class labels precisely.

\subsubsection{Influence of feature alignment}
The alignment ablation studies on the~\ourModel\ model aim to assess its effectiveness. We compare aligning the OCR with image features alone (C2I) or text features alone (C2T) against aligning with both modalities. The results, as detailed in \textit{Tab.~\ref{Alignment-ablation-studies-RVL-CDIP}}, show varying degrees of performance across the four sequential splits (A to D) for the RVL-CDIP dataset. Within the ZSL setting, the \ourModel~model's performance does not favor a single alignment strategy; both C2T and C2I alignments prove crucial across various splits. Specifically, C2T alignment is vital for splits A and B, whereas C2I alignment is key for splits C and D. For GZSL, the combined alignment of \ourModel~consistently outperforms the singular alignment models. This affirms the importance of multimodal integration for enhanced model performance. Overall, the combined alignment approach of the \ourModel~model demonstrates the most balanced performance, particularly in GZSL scenarios, where it achieves higher harmonic mean (\textbf{H}), showcasing its superior capability to generalize across both seen and unseen classes.

\begin{table}
    \centering
    \caption{Performance outcomes of \ourModel~on sequential splits (A - D) of the RVL-CDIP dataset, demonstrating alignment ablations: 'C2I' for sole Content-to-Image alignment and 'C2T' for sole Content-to-Text alignment. We observe that aligning content to both text and image improves the performance of \ourModel. Best values are \textbf{bold}.}
    \begin{tabular}{clcccc}
    \toprule
    & & \multicolumn{4}{c}{\textbf{RVL-CDIP}} \\
    \cmidrule(l){3-6}
    & & \textbf{ZSL} & \multicolumn{3}{c}{\textbf{GZSL}} \\
    \cmidrule(l){3-3} \cmidrule(l){4-6}
    \textbf{Split} & \textbf{Method} & \textbf{T1} & \textbf{u} & \textbf{s} & \textbf{H} \\
    \cmidrule(l){1-1} \cmidrule(l){2-2} \cmidrule(l){3-3} \cmidrule(l){4-4} \cmidrule(l){5-5} \cmidrule(l){6-6}
    \multirow{2}{*}{A}
    & \texttt{\ourModel~(C2I) } & 68.25 & 55.21 & 49.85 & 52.39 \\
    & \texttt{\ourModel~(C2T)} & 68.85 & 57.37 & 69.56 & 62.88 \\
    & \texttt{\ourModel } & \textbf{70.64} & 61.84 & 69.36 & \textbf{65.38} \\
    \midrule
    \multirow{2}{*}{B}
    & \texttt{\ourModel~(C2I)} & 75.63 & 65.25 & 50.80 & 57.13 \\
    & \texttt{\ourModel~(C2T)} & 77.26 & 65.13 & 75.28 & 69.84 \\
    & \texttt{\ourModel } & \textbf{77.87} & 69.84 & 75.40 & \textbf{72.52} \\
    \midrule
    \multirow{2}{*}{C}
    & \texttt{\ourModel~(C2I)} & \textbf{63.94} & 55.94 & 50.79 & 53.24 \\
    & \texttt{\ourModel~(C2T)} & 51.65 & 45.76 & 69.55 & 55.21 \\
    & \texttt{\ourModel } & 59.52 & 53.94 & 72.17 & \textbf{61.74} \\
    \midrule
    \multirow{2}{*}{D}
    & \texttt{\ourModel~(C2I)} & 59.28 & 50.21 & 46.47 & 48.27 \\
    & \texttt{\ourModel~(C2T)} & 54.54 & 45.92 & 66.47 & 54.31 \\
    & \texttt{\ourModel } & \textbf{61.13} & 51.13 & 71.17 & \textbf{59.51} \\
    \bottomrule
    \end{tabular}
    \label{Alignment-ablation-studies-RVL-CDIP}
\end{table}


\subsubsection{Influence of OCR engine}
In our OCR ablation studies for the \ourModel~model, we examine the impact of different OCR engines on the model's efficacy. We compare Tesseract OCR with the default docTR OCR engine. \textit{Tab.~\ref{OCR-ablation-studies-RVL-CDIP}} shows our findings. In the ZSL setting, the performance boost with docTR is not uniform across all evaluated splits. This is evident from the top-1 accuracy rates. The situation changes in the GZSL setting. Here, docTR significantly enhances accuracies for both unseen and seen classes. It also achieves a better harmonic mean (\textbf{H}). These results highlight the crucial role of OCR quality. It improves model accuracy for both familiar and new classes. docTR's accuracy in providing OCR greatly benefits \ourModel~'s predictive performance.

\begin{table}[H]
    \centering
    \caption{Performance with OCR engine variations on sequential splits (A - D) of RVL-CDIP. We observe \ourModel's sensitivity to OCR quality. Best values are \textbf{bold}.}
    \begin{tabular}{clcccc}
    \toprule
    & & \multicolumn{4}{c}{\textbf{RVL-CDIP}} \\
    \cmidrule(l){3-6}
    & & \textbf{ZSL} & \multicolumn{3}{c}{\textbf{GZSL}} \\
    \cmidrule(l){3-3} \cmidrule(l){4-6}
    \textbf{Split} & \textbf{Method} & \textbf{T1} & \textbf{u} & \textbf{s} & \textbf{H} \\
    \cmidrule(l){1-1} \cmidrule(l){2-2} \cmidrule(l){3-3} \cmidrule(l){4-4} \cmidrule(l){5-5} \cmidrule(l){6-6}
    \multirow{2}{*}[0.5em]{A}
    & \texttt{\ourModel~(Tesseract)} & 64.10 & 54.70 & 51.53 & 53.07 \\
    & \texttt{\ourModel~(docTR)} & \textbf{70.64} & 61.84 & 69.36 & \textbf{65.38} \\
    \midrule
    \multirow{2}{*}[0.5em]{B}
    & \texttt{\ourModel~(Tesseract)} & 67.88 & 57.13 & 59.61 & 58.35 \\
    & \texttt{\ourModel~(docTR)} & \textbf{77.87} & 69.84 & 75.40 & \textbf{72.52} \\
    \midrule
    \multirow{2}{*}[0.5em]{C}
    & \texttt{\ourModel~(Tesseract)} & \textbf{66.77} & 52.12 & 57.92 & 54.87 \\
    & \texttt{\ourModel~(docTR)} & 59.52 & 53.94 & 72.17 & \textbf{61.74} \\
    \midrule
    \multirow{2}{*}[0.5em]{D}
    & \texttt{\ourModel~(Tesseract)}& \textbf{65.48} & 52.17 & 51.38 & 51.77 \\
    & \texttt{\ourModel~(docTR)} & 61.13 & 51.13 & 71.17 & \textbf{59.51} \\
    \bottomrule
    \end{tabular}
    \label{OCR-ablation-studies-RVL-CDIP}
\end{table}


\subsubsection{Influence of inference strategy}
The inference ablation studies for \ourModel~examine the effects of different fusion techniques on predicting class labels. \textit{Tab.~\ref{Inference-ablation-studies-RVL-CDIP}} outlines these comparisons. The standard method employs late fusion, first integrating content encoding $C_1$ and image encoding $I_1$ with text encoding $T_1$. Subsequently, it calculates the mean of these dot products, as depicted in \textit{Fig. \ref{Content-CLIP-Inference}}. Conversely, the early fusion method merges $I_1$ and $C_1$ before the dot product with $T_1$. Results indicate that \ourModel's late fusion performs comparably to early fusion in ZSL scenarios, with minor differences observed in splits 'B' and 'C'. In GZSL scenarios, however, late fusion surpasses early fusion across all splits. This suggests that processing image and content data separately retains detailed features, improving accuracy. The harmonic mean (\textbf{H}) is consistently higher without early fusion. This demonstrates that this approach better supports generalization across both seen and unseen classes.

\begin{table}[H]
    \centering
    \caption{\ourModel's inference fusion ablation outcomes across sequential splits (A - D) of RVL-CDIP. Early fusion leads to poor GZSL performance. Best values are \textbf{bold}.}
    \begin{tabular}{clcccc}
    \toprule
    & & \multicolumn{4}{c}{\textbf{RVL-CDIP}} \\
    \cmidrule(l){3-6}
    & & \textbf{ZSL} & \multicolumn{3}{c}{\textbf{GZSL}} \\
    \cmidrule(l){3-3} \cmidrule(l){4-6}
    \textbf{Split} & \textbf{Method} & \textbf{T1} & \textbf{u} & \textbf{s} & \textbf{H} \\
    \cmidrule(l){1-1} \cmidrule(l){2-2} \cmidrule(l){3-3} \cmidrule(l){4-4} \cmidrule(l){5-5} \cmidrule(l){6-6}
    \multirow{2}{*}[0.5em]{A}
    & \texttt{\ourModel~(Early Fusion)} & 70.63 & 65.86 & 59.22 & 62.36 \\
    & \texttt{\ourModel } & \textbf{70.64} & 61.84 & 69.36 & \textbf{65.38} \\
    \midrule
    \multirow{2}{*}[0.5em]{B}
    & \texttt{\ourModel~(Early Fusion)} & \textbf{77.88} & 76.34 & 42.01 & 54.20 \\
    & \texttt{\ourModel } & 77.87 & 69.84 & 75.40 & \textbf{72.52} \\
    \midrule
    \multirow{2}{*}[0.5em]{C}
    & \texttt{\ourModel~(Early Fusion)} & \textbf{59.55} & 59.01 & 45.25 & 51.22 \\
    & \texttt{\ourModel } & 59.52 & 53.94 & 72.17 & \textbf{61.74} \\
    \midrule
    \multirow{2}{*}[0.5em]{D}
    & \texttt{\ourModel~(Early Fusion)}& 61.11 & 56.32 & 53.42 & 54.83 \\
    & \texttt{\ourModel } & \textbf{61.13} & 51.13 & 71.17 & \textbf{59.51} \\
    \bottomrule
    \end{tabular}
    \label{Inference-ablation-studies-RVL-CDIP}
\end{table}

\section{Conclusion}

In our study, we introduce \ourModel, a novel approach for zero-shot learning in document image classification. We establish dataset splits to provide a baseline for future research and incorporate a 'content module' for leveraging OCR data. This approach can also integrate other textual information, such as metadata. Our innovative 'coupled contrastive' loss mechanism aligns the features of the content module with those of CLIP, enhancing the model's performance. Specifically, our method improves CLIP's ZSL top-1 accuracy by 6.7\% and GZSL harmonic mean by 24\% on the RVL-CDIP dataset. These improvements, coupled with our development of specific ZSL and GZSL dataset splits, highlight \ourModel's effectiveness and potential to advance document management systems by improving model generalization across unseen document types.
\bibliographystyle{splncs04}
\bibliography{references}
\end{document}